\documentclass{svproc}
\usepackage{url}

\usepackage[utf8]{inputenc}
\usepackage{amsmath,amssymb,amstext,amsxtra,amsbsy,amsgen,accents}
\usepackage{graphicx}
\usepackage{pgfgantt}
\usepackage{pdfpages}
\usepackage{longtable}
\usepackage{multirow}
\usepackage{makecell}
\usepackage{algpseudocode,algorithm}
\usepackage{rotating} 
\usepackage{enumitem}
\usepackage{subcaption}

\newcommand{\mcite}[1]{\cite{#1}\textrm{\color{magenta}}}
\newcommand{\mref}[1]{\ref{#1}\textrm{\color{magenta}}}
\newcommand{\mlabel}[1]{\label{#1}\textrm{\color{magenta}}}

\newcommand{\ff}{\ensuremath{\boldsymbol{O}\text{-}\bX\bY\bZ}\xspace}
\newcommand{\mf}{\ensuremath{\boldsymbol{o}\text{-}\bx\by\bz}\xspace}
\newcommand{\bbi}{\ensuremath{\bb_i}\xspace}
\newcommand{\bti}{\ensuremath{\bt_i}\xspace}
\newcommand{\bai}{\ensuremath{\ba_i}\xspace}

\newcommand{\sphrf}{\ensuremath{\mathcal{S}'}\xspace}

\newcommand{\tilminri}{\ensuremath{\undertilde{r}}\xspace}
\newcommand{\tilmaxri}{\ensuremath{\widetilde{r}}\xspace}

\newcommand{\gmb}{\ensuremath{\gamma_\text{f}}\xspace}

\newcommand{\gmt}{\ensuremath{\gamma_\text{m}}\xspace}

\newcommand{\rt}{\ensuremath{r_\text{m}}\xspace}
\newcommand{\rbb}{\ensuremath{r_\text{f}}\xspace}
\newcommand{\rthr}{\ensuremath{\mathbb{R}^3}\xspace}

\newcommand{\nso}{\ensuremath{N_\text{S}}\xspace}

\newcommand{\nr}{\ensuremath{N_\text{r}}\xspace}
\newcommand{\ns}{\ensuremath{N_\text{s}}\xspace}

\newcommand{\rc}{\ensuremath{r_\text{c}}\xspace}
\newcommand{\cyi}{\ensuremath{\mathcal{C}_i}\xspace}
\newcommand{\cyj}{\ensuremath{\mathcal{C}_j}\xspace}
\newcommand{\dij}{\ensuremath{d_{i,j}}\xspace}
\newcommand{\sthree}{\ensuremath{\mathcal{S}_3}\xspace}

\renewcommand\appendix{\par
	\setcounter{section}{0}
	\setcounter{subsection}{0}
	\setcounter{figure}{0}
	\setcounter{table}{0}
	\renewcommand\thesection{Appendix \Alph{section}}
	\renewcommand\thefigure{\Alph{section}\arabic{figure}}
	\renewcommand\thetable{\Alph{section}\arabic{table}}
}

\usepackage{xspace}
\usepackage{amsmath,amssymb,amsfonts}

\newcommand{\bs}[1]{\ensuremath{\boldsymbol{#1}}}

\newcommand{\ba}{\ensuremath{\bs a}\xspace}
\newcommand{\bb}{\ensuremath{\bs b}\xspace}
\newcommand{\bc}{\ensuremath{\bs c}\xspace}

\newcommand{\bo}{\ensuremath{\bs o}\xspace}
\newcommand{\bp}{\ensuremath{\bs p}\xspace}

\newcommand{\bt}{\ensuremath{\bs t}\xspace}

\newcommand{\bx}{\ensuremath{\bs x}\xspace}
\newcommand{\by}{\ensuremath{\bs y}\xspace}
\newcommand{\bz}{\ensuremath{\bs z}\xspace}

\newcommand{\bO}{\ensuremath{\bs O}\xspace}

\newcommand{\bR}{\ensuremath{\bs R}\xspace}

\newcommand{\bX}{\ensuremath{\bs X}\xspace}
\newcommand{\bY}{\ensuremath{\bs Y}\xspace}
\newcommand{\bZ}{\ensuremath{\bs Z}\xspace}

\newcommand{\Del}{\ensuremath{\Delta}\xspace}

\renewcommand{\Re}{\ensuremath{\mathbb{R}}\xspace} 

\newcommand{\SE}{\ensuremath{\mathbb{SE}(3)}\xspace}

\newcommand{\SO}{\ensuremath{\mathbb{SO}(3)}\xspace}

\newcommand{\dofs}{degrees-of-freedom\xspace}

\providecommand{\keywords}[1]
{
	\small	
	\textbf{\textit{Keywords---}} #1
}

\newcommand{\srs}{SRSPM\xspace}
\newcommand{\spm}{SGPM\xspace}
\newcommand{\sps}{6-S\underline{P}S\xspace}
\newcommand{\cfs}{CFS\xspace}
\newcommand{\api}{API\xspace}
\newcommand{\invent}{\textsf{Autodesk Inventor}\xspace}

\begin{document}
	%
	\title{Validation of collision-free spheres of Stewart-Gough platforms for constant orientations using the Application Programming Interface of a CAD software}
	\titlerunning{Validation of the CFSs of Stewart-Gough platforms using a CAD software}  
	%
	\author{Bibekananda Patra\inst{1}, Rajeevlochana G. Chittawadigi\inst{2}, and Sandipan Bandyopadhyay\inst{1}}
	\date{}
	\authorrunning{B. Patra et al.} 
	%
	%
	\institute{Indian Institute of Technology Madras, Chennai 600036, India,\\
		\email{bibeka.patra2@gmail.com, sandipan@iitm.ac.in},\\
		WWW home page:
		\texttt{https://ed.iitm.ac.in/$\sim$sandipan/}
		\and
		Amrita Vishwa Vidyapeetham, Bengaluru, 560035, India,\\ 
		\email{rg\_chittawadigi@blr.amrita.edu},\\
		WWW home page:
		\texttt{https://www.amrita.edu/faculty/rg-chittawadigi/}
		}
	
	\maketitle              
	
	\begin{abstract}
		This paper presents a method of validation of the size of the largest collision-free sphere~(\cfs) of a $6$-$6$ Stewart-Gough platform manipulator~(\spm) for a given orientation of its moving platform (MP) using the Application Programming Interface~(\api) of a CAD software. The position of the MP is updated via the~\api in an automated manner over a set of samples within a shell enclosing the surface of the CFS. For each pose of the manipulator, each pair of legs is investigated for mutual collisions. The CFS is considered safe or validated iff none of the points falling inside the CFS lead to a collision between any pair of legs. 
		This approach can not only validate the safety of a precomputed~\cfs, but also estimate the same for any spatial parallel manipulator.

	\end{abstract}
	\section{Introduction}
	\mlabel{sc:intro}
	Parallel manipulators (PMs) have one or more {\em passive} (i.e., un-actuated) links in their architecture, in addition to actuated ones. The greater number of links in PMs increases the chances of collisions among the links during their operation, which can be detrimental to their surroundings as well as themselves. One way to circumvent this issue is to determine \emph{a priori} a subset of its workspace that is free of collisions among links, and confine all planned motions of the manipulator within this space. 
	
	In general, links of PMs have irregular geometries, rendering the detection of collisions among them computationally challenging. Therefore, in practice, these links are approximated by convex envelopes, so as to reduce the associated computational complexity. 
	For instance, cylinders were used to represent the links in certain spatial PMs~\mcite{Majid2000},~\mcite{Oen2006},~\mcite{NagS32019}. In~\mcite{Dhruvesh2017}, cuboidal bounding boxes were used for the same purpose. 
	
	To compute a \emph{region} free of collisions among the links of a spatial PM, its workspace, which is a subset of~\SE, may be notionally decomposed into its constituent subgroups, namely,~\SO and~\rthr, as described in~\mcite{PatraS32024}. While the orientation workspace is approximated by a set of discrete samples inside it (each of which corresponds to a distinct orientation of the moving platform (MP)), the determination of the collision-free region in~\rthr is rendered amenable to analytical computation by modelling it as a sphere, termed the \emph{collision-free sphere}~(\cfs)~\mcite{PatraS32024}. 

	In this paper, the focus is on the development of a computational tool to numerically validate a pre-computed~CFS of a  \emph{semi-regular Stewart-Gough platform manipulator}~(\srs), which was studied in~\mcite{PatraS32024}. In this context,  
	``validation''\footnote{In this validation, the limits on leg extensions, the occurrence of the \emph{gain-type singularities}, and passive joint limit violations are not investigated.} implies a computational verification of the claim that as long as the centre of the MP lies inside the~\cfs corresponding to a given orientation of the MP, {\em none} of the pairs of legs of the manipulator can collide. The motivation behind creating such a tool is to attain the ability to verify the results obtained via the analytical approach presented in~\mcite{PatraS32024}. Even though the analytical results are expected to be superior in terms of accuracy, they are not immune to accidental errors due to human interventions, say, in the process of specifying geometric data. An independent validation is deemed necessary, given the potentially grim consequences of an actual collision during operation. Obviously, such a validation cannot be of a physical nature, for the same set of risks stated above. 
	
	In this work, the task of geometric computations is relegated to a commercial CAD software, namely,~\textsf{Autodesk Inventor\footnote{The software~\textsf{Autodesk Inventor Professional 2023}  is referred to as~\invent hereinafter for brevity.} Professional 2023}~\mcite{Autodesk}. While it is possible to perform these calculations analytically, as shown in~\mcite{NagS32019}, the process described therein involves the derivation and solution of a polynomial of degree~8, for each pair of legs, for each position of the MP. Obviously, such computations are complicated, as well as demanding, and may not be easy to implement. An improved yet simplified method for computing the~\cfs is presented in~\mcite{PatraS32024}, where \emph{capsules} have been used to approximate the legs, as opposed to cylinders in~\mcite{NagS32019}. This work, too, uses capsules to approximate the legs. However, it leverages the geometric computation engine of ~\textsf{Autodesk Inventor} via its Application Programming Interface~(API). Thus, fast and reliable development of the validation tool is rendered feasible. Such applications in the domain of collision detection among the rigid bodies, etc., have been discussed in detail in~\mcite{Rajeev2024} and the references therein. However, to the best of the authors' knowledge, an application of such an idea to a 6-\dofs spatial parallel manipulator is entirely novel, and it is hoped that it may inspire similar applications in the future, rendering the subject of parallel manipulators more accessible to analysts. 
	
	The rest of the paper is organised as follows. Section~\mref{sc:math} describes the geometry of the~\srs, followed by a brief discussion on the computation of the~\cfs for a given orientation of the MP. Section~\mref{sc:validapi} discusses the method of validating the~\cfs via an API of~\invent. The method is implemented for a given architecture of the~\srs in Section~\mref{sc:result}. Finally, the article is concluded in Section~\mref{sc:concl}.
	\section{Mathematical preliminaries}
	\mlabel{sc:math}
	This section discusses the geometry of the~\srs, followed by the  computation of the CFS for a fixed orientation of the MP.
	
	\subsection{Geometry of an SRSPM}
	\mlabel{sc:geomsrspm}
	\begin{figure}[t]
		\centering
		\begin{minipage}{0.45\textwidth}
			\centering
			\includegraphics[width=\textwidth]{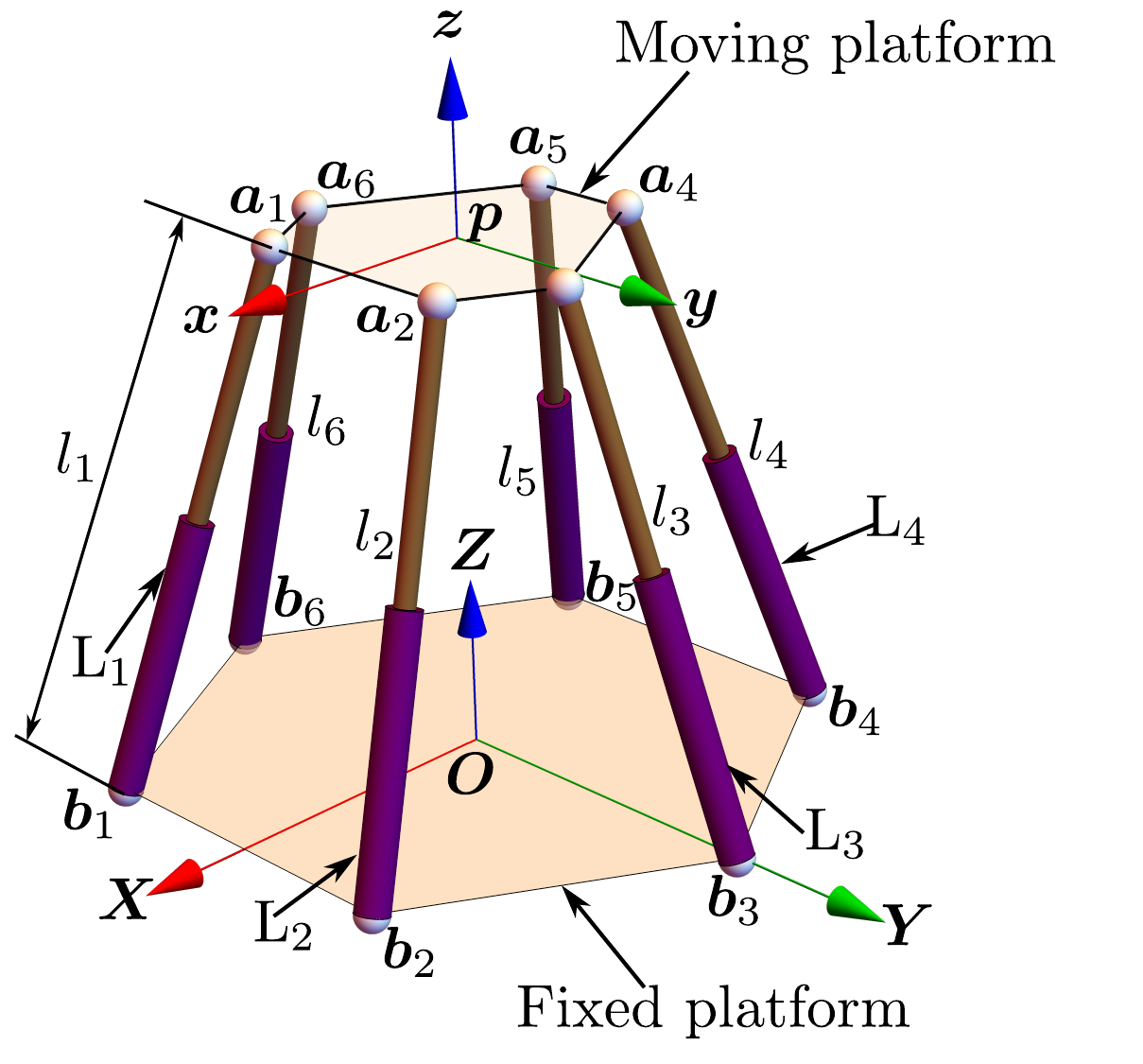}
			\caption{Schematic of an~\srs. The vertices of the fixed and moving platforms,~\bbi and~\bai, respectively, are represented in~\ff.}
			\mlabel{fg:srspm}
		\end{minipage}
		\hfill
		\begin{minipage}{0.45\textwidth}
			\centering
			\includegraphics[width=\textwidth]{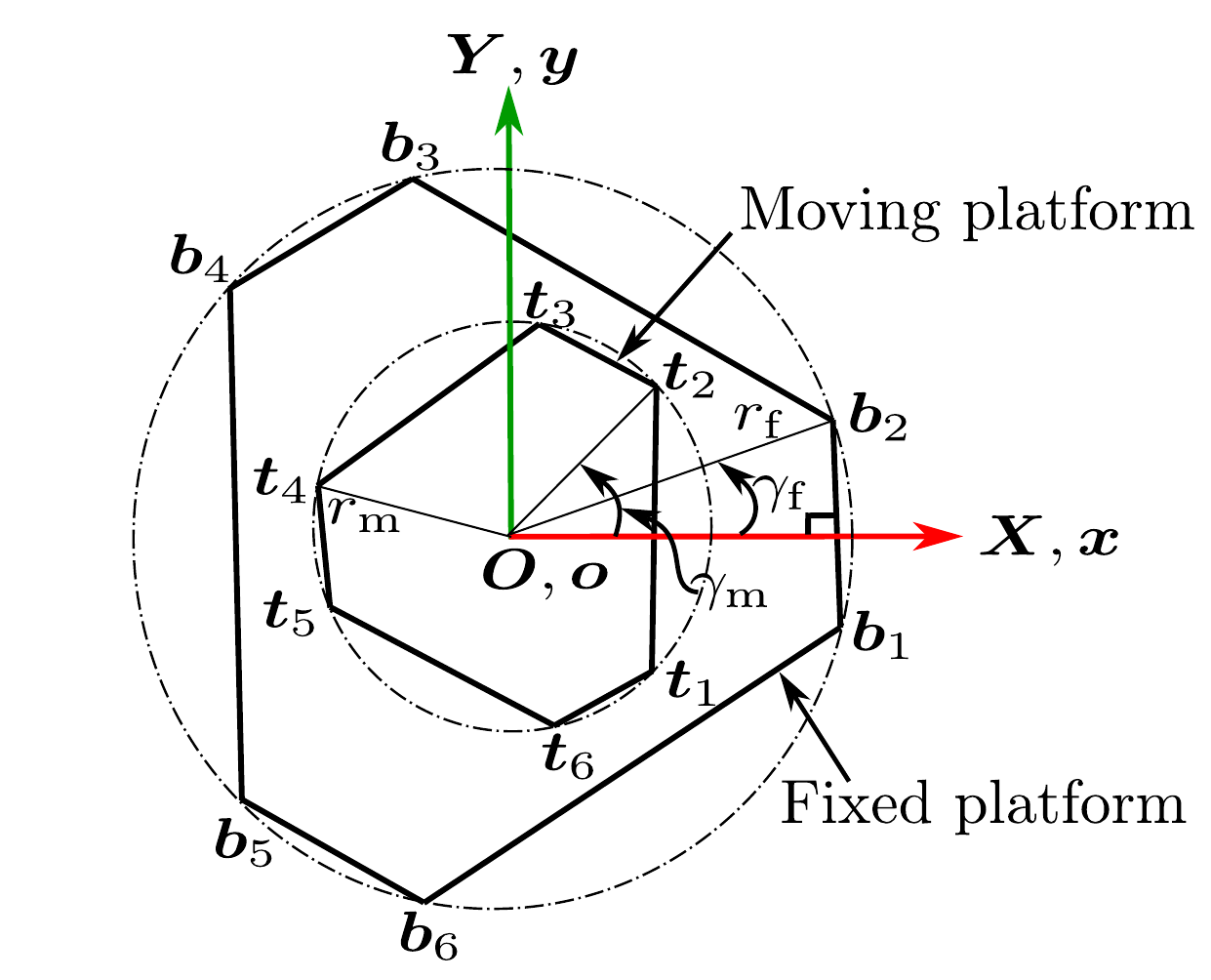}
			\caption{Geometry of the fixed and moving platforms. The vertices of the fixed and moving platforms,~\bbi and~\bti, are represented in~\ff and~\mf, respectively.}
			\mlabel{fg:platforms}
		\end{minipage}
	\end{figure}
	The schematic of an~\srs is shown in Fig.~\mref{fg:srspm}. This manipulator has the~\sps kinematic architecture; i.e., it is mobilised by six linear actuators, which are connected via spherical joints (un-actuated) mounted on the fixed platform (FP), and the MP. The centres of the spherical joints are located at the vertices of both platforms. Figure~\mref{fg:platforms} depicts the geometry of both FP and MP. The circum-radii of the FP and the MP are denoted by~\rbb and~\rt, respectively. 
	The fixed and moving frames of reference, namely,~\ff and~\mf, are attached to the FP and the MP, respectively, with their origins~\bO and~\bo, located at the centroids of the corresponding platforms, as shown in Fig.~\mref{fg:platforms}. The angular displacements\footnote{All the angles are in radians, unless mentioned otherwise.} of the spherical joints mounted on the FP, measured from the positive~\bX-axis in a CCW manner, are as follows: $\left\{-\gmb, \gmb, \frac{2\pi}{3}-\gmb, \frac{2\pi}{3}+\gmb, \frac{4\pi}{3}-\gmb, \frac{4\pi}{3}+\gmb\right\}$, where~$2\gmb$ is the angle subtended by the two vertices immediate adjacent to the~\bX-axis at the centroid of the platform. Similarly, the angular displacements of the spherical joints mounted on the MP, measured from the positive~\bx-axis in a CCW manner, are:~$\left\{-\gmt, \gmt, \frac{2\pi}{3}-\gmt, \frac{2\pi}{3}+\gmt, \frac{4\pi}{3}-\gmt, \frac{4\pi}{3}+\gmt\right\}$, where~$2\gmt$ is defined in an analogous manner. The vertices of the FP and the MP are denoted by~\bbi, and~\bti,~$i = 1,\dots,6$, in their respective frames of reference~\ff, and~\mf, respectively. The origin~$\bo$ of~\mf is mapped to the point~$\bp=[x,y,z]^\top$ when represented in the frame~\ff (see Fig.~\mref{fg:srspm}). Similarly, the vertices~\bti of the MP are mapped to the points~\bai in~\ff as follows:~$\bai = \bp + \bR \bti,$ where~$\bR \in \SO$ is the matrix representing the orientation of~\mf w.r.t.~\ff. The matrix~\bR is expressed in terms of \emph{Rodrigues parameters},~$\bc = [c_1,c_2,c_3]^\top \in \Re^3$ (see~\mcite{PatraSO32025} and the references therein for details). 
	
	\subsection{Collision-free sphere for a constant orientation of the MP}
	\mlabel{sc:erwconst}
	\begin{figure}[t]
		\centering
		\begin{minipage}{\textwidth}
			\centering
			\includegraphics[width=0.6\textwidth]{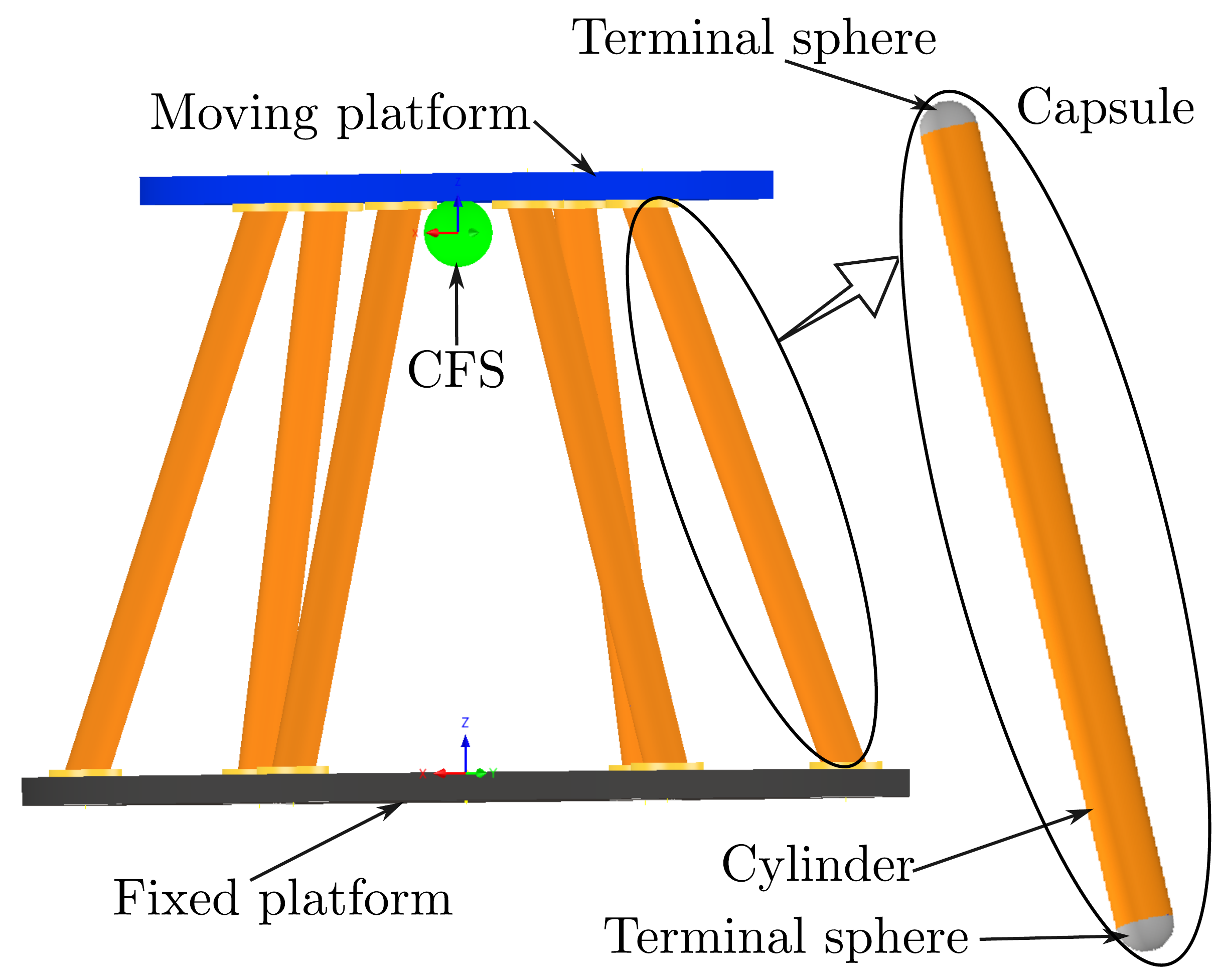}
			\caption{A geometrically equivalent CAD model of an \srs (refer to Table~\mref{tb:archparam} for the architecture details), showing the \cfs along with a zoomed view of a capsule}
			\mlabel{fg:cadsrs}
		\end{minipage}
	\end{figure}
	The derivation of the \cfs\footnote{In this work, collisions among the legs are studied, since in practice, collisions between the legs and the two platforms typically do not occur due to the constraints imposed by both active and passive joint limits.} for a given orientation of the MP has been presented in~\mcite{PatraS32024} in detail. This has been discussed briefly below for the sake of completeness. 
	
	As mentioned in Section~\mref{sc:geomsrspm}, the legs of the manipulator consist of linear actuators and the ball parts of the  spherical joints at both ends. Typical mechanical construction of the actuators makes their geometry complicated and  irregular to some extent. To simplify the mathematical formulation for collision detection among the legs, each leg is enclosed within a \emph{capsule}~\mcite{Safeea2019}, consisting of a cylinder and two terminal spheres (see Fig.~\mref{fg:cadsrs}), each having a uniform radius, denoted by~\rc.
	Therefore, the problem of detecting collisions between a pair of legs reduces to the identification of collisions among the corresponding capsules. This is achieved by investigating if two capsules, denoted by~\cyi and~\cyj,~\mbox{$i \neq j$}, are \emph{tangent} to each other. Given that the shortest distance between two capsules may be computed by deducting twice the radius~\rc from the perpendicular distance (denoted by~\dij) between their axes, the condition of the tangency between~\cyi and~\cyj reduces to~$S_3:=\dij^2 - 4\rc^2 = 0$, which defines a \emph{collision surface} that is found to be a quadratic in~$\{x, y, z\}$.

	Next, the largest CFS, denoted by~\sthree, needs to be determined in such a manner that inside~\sthree, the legs of the manipulator cannot interfere with each other. The centre (denoted by~$\bp_0 = [0,0,z_0]^\top$, where~$z_0$ is known as the \emph{neutral height}~\mcite{Nabavi2018} of the MP) of~\sthree is on the~\bZ-axis perpendicular to the FP and passing through the origin~\bO. The sphere~\sthree is computed in such a way that it is tangent to the collision surface~$S_3=0$, without intersecting the surface at any other point. 
	
	\section{Method of validation of a CFS using the API}
	\mlabel{sc:validapi}
	This section discusses the method of validating the \cfs for a fixed orientation of the MP using an \emph{add-in} utilising the~\api of~\invent.
	\subsection{Developing an add-in using the~\api of~\invent}
	\mlabel{sc:api}
	\begin{figure}[t]
		\begin{minipage}{\textwidth}
			\centering
			\includegraphics[width=0.8\textwidth]{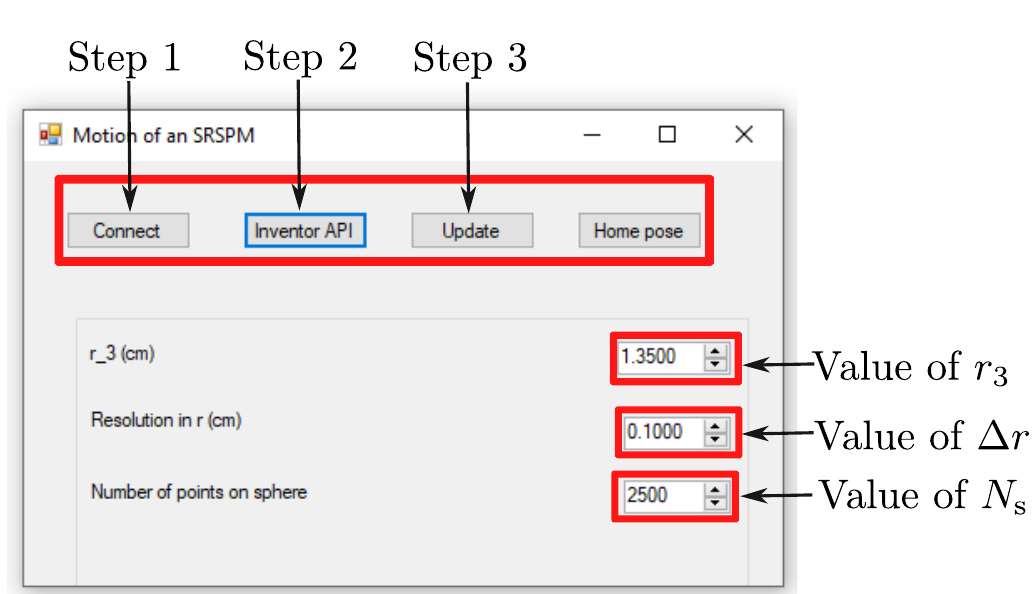}
			\caption{An add-in to \invent for validating the~\cfs of the~\srs. The values~$r_3, \Del r,$ and~\ns shown here correspond to Scenario~2 (see Tables~\ref{tb:testcase},~\ref{tb:paramscan}) }
			\label{fg:gui}
		\end{minipage}
	\end{figure}
	\begin{table}[t]
		\centering
		\begin{minipage}{0.45\textwidth}
			\centering
			\caption{Architecture parameters of the~\srs and its neutral height}
			\mlabel{tb:archparam}
			\begin{tabular}{|l|l|r|}
				\hline
				\makecell{Significance} & Symbol & \makecell{Value} \\
				\hline
				\makecell[l]{Circum-radius of \\the FP} & \rbb & 150.0 mm  \\
				\hline
				\makecell[l]{Circum-radius of\\ the MP} & \rt & 75.0 mm \\
				\hline
				\makecell[l]{Half of the angle \\subtended by~$\bb_1$ and\\ $\bb_2$ at the centre (\bO) \\of the FP} & \gmb & $30.5^\circ$ \\
				\hline
				\makecell[l]{Half of the angle \\subtended by~$\bt_1$ and\\ $\bt_2$ at the centre (\bo)\\of the MP} & \gmt & $40.5^\circ$ \\
				\hline 
				\makecell[l]{Radius of the \\ enclosing capsule} & \rc & 8.5 mm \\
				\hline
				Neutral height & $z_0$ & 300.0 mm\\
				\hline
			\end{tabular}
		\end{minipage}
		\hfill
		\begin{minipage}{0.45\textwidth}
			\centering
			\setlength\extrarowheight{4pt}
			\caption{Orientation of the moving platform and the corresponding radius of the~\cfs for two scenarios (refer to~\cite{PatraS32024} for details)}
			\mlabel{tb:testcase}
			\begin{tabular}{|c|r|r|}
				\hline
				 Scenario & \makecell{Rodrigues parameters~(\bc)} & \makecell{$r_3$ (mm)} \\
				\hline
				1 & $[-0.2301, 0.0413, 3.0209]^\top$ & $188.4$\\
				\hline
				2 & $[0.2534, 0.6740, 0.2653]^\top$ & $13.5$\\
				\hline
			\end{tabular}
			\bigskip{}
			\caption{Controlling parameters for generating samples in the shell~\sphrf}
			\mlabel{tb:paramscan}
			\begin{tabular}{|l|l|r|r|}
				\hline
				\multirow{2}{*}{Significance} & \multirow{2}{*}{Symbol} & \multicolumn{2}{|c|}{Value}\\
				\cline{3-4}
				& & Scenario~1 & Scenario 2\\
				\hline
				\makecell[l]{Inner radius\\ of~\sphrf} & \tilminri & 169.6 mm & 12.2 mm\\
				\hline
				\makecell[l]{Outer radius\\ of~\sphrf} & \tilmaxri & 207.2 mm & 14.9 mm\\
				\hline
				\makecell[l]{Resolution in~$r$} & $\Del r$ & 10 mm & 1 mm\\
				\hline
				\makecell[l]{Number of \\points in~\sphrf} & $N$ & 10,000 & 7,500\\
				\hline
			\end{tabular}
		\end{minipage}
	\end{table}
	The functional details of the \api for \invent can be found in~\mcite{Rajeev2024}. These have been summarised below for the sake of completeness.
	
	\invent is a commercially available CAD software that enables its users to access parts and assemblies defined in its computational environment through add-ins developed in the~\textsf{Visual C\#} language, running on the \textsf{Windows} operating system. Using this functionality, one can automate one or both of the following tasks: (a)~editing the geometry of the parts of the assemblies, and (b)~manipulating objects virtually -- i.e., modifying their positions and/or orientations. The second capability is utilised in this work. 
	
	To begin with, the model of an~\srs has been created in~\invent, as shown in Fig.~\mref{fg:cadsrs}. As may be seen therein, the legs of the manipulator are enclosed within their respective capsules.
	Next, an add-in for~\invent is developed using the API, as shown in Fig.~\mref{fg:gui}, which has four features, namely, ``\verb|Connect|'', ``\verb|Inventor API|'', ``\verb|Update|'', and ``\verb|Home pose|''. While the first three features are used to initiate the validation, as discussed below, the fourth feature is used to reset the manipulator to its neutral position,~$\bp_0$,  before starting and as well as after completing the validation process. 
	The following steps are used to validate the \cfs using the first three features of the add-in.
	\begin{enumerate}
		\item[\textbf{Step 1}:] \textbf{Connect}: This feature checks if~\invent is active. If not, the add-in opens a dialogue box asking the user to open the application.   
		\item[\textbf{Step 2}:]\textbf{Inventor API}: Once the connection is established, this feature checks whether the file opened contains a valid assembly. If it does not, it prompts the user to open another with a valid assembly in it. For a valid file, it extracts all the data defining the model, including the geometry of the parts, constituents of each sub-assembly, kinematics constraints between parts, etc. 
		\item[\textbf{Step 3}:]\textbf{Update}: Once the API has acquired all the information of the CAD model, the ``\verb|Update|'' feature sends the position information of the point~\bp, which is the centroid of the MP (see Fig.~\mref{fg:srspm}), from a \emph{user-defined function} of the~\api to the computation engine of the~\invent. The generation of such positions is explained in Section~\mref{sc:validcfs}. Since the kinematic constraints of the SRSPM are incorporated entirely in~\invent, updates in the position of the MP automatically impose the corresponding changes in the positions and orientations of the legs of the manipulator.
		\item[\textbf{Step 4}:] For a given orientation and each of the  positions of the MP specified in Step~3, each pair of legs is queried for collisions among themselves.  
	\end{enumerate}  
	Steps~1-3 are depicted in Fig.~\mref{fg:gui}, whereas Step~4 is discussed in the following. 
	\subsection{Validation of the collision-free sphere}
	\mlabel{sc:validcfs}
	\begin{figure}[t]
		\centering
		\begin{subfigure}{0.45\textwidth}
			\centering
			\includegraphics[width=\textwidth]{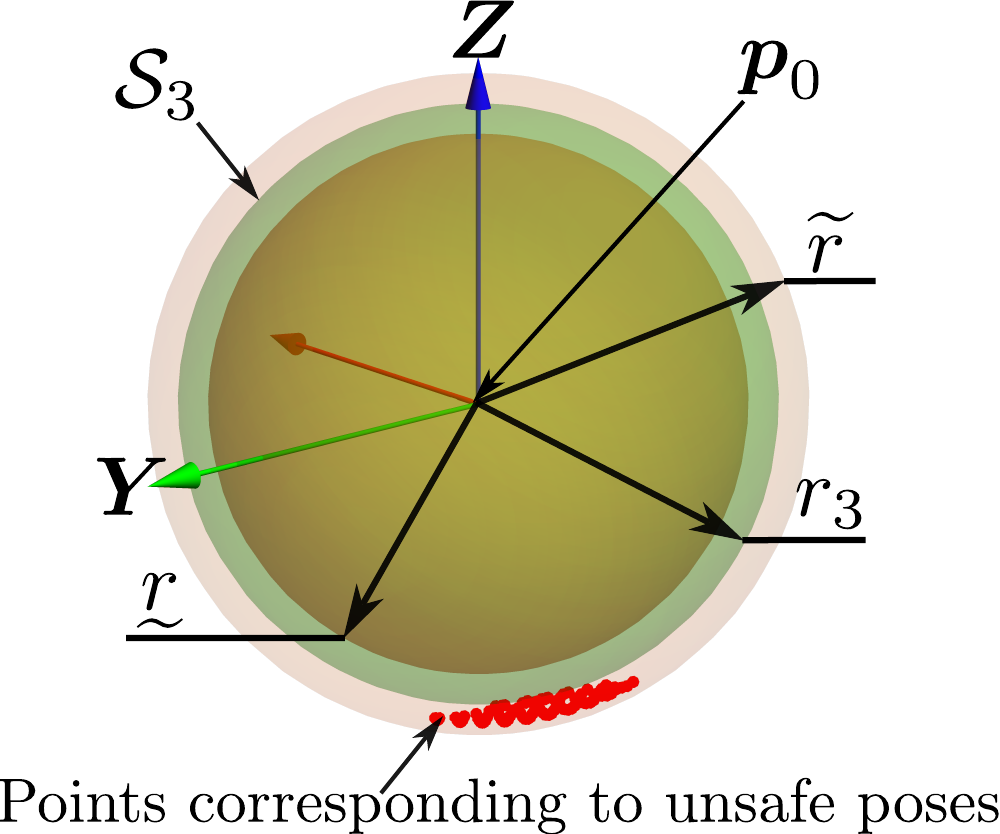}
			\caption{Scenario~1}
			\mlabel{fg:unsafecase1}
		\end{subfigure}
		\hfill
		\begin{subfigure}{0.45\textwidth}
			\centering
			\includegraphics[width=\textwidth]{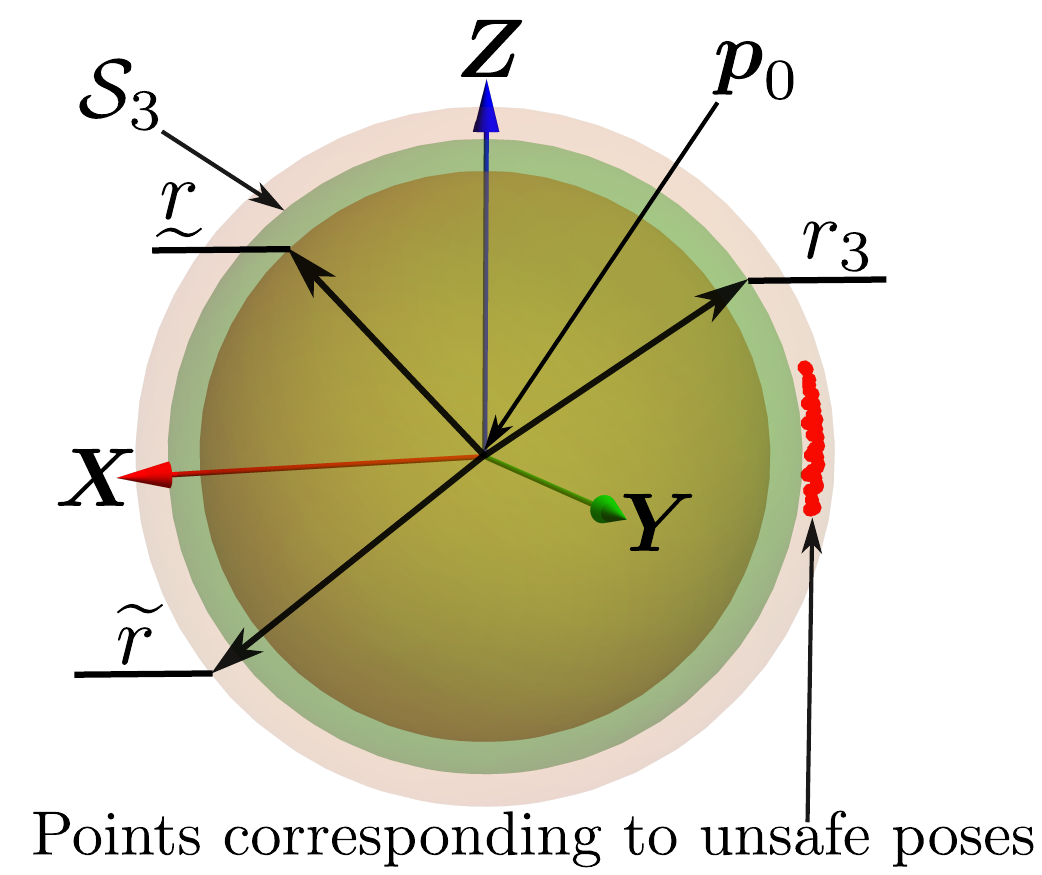}
			\caption{Scenario~2}
			\mlabel{fg:unsafecase2}
		\end{subfigure}
		\caption{Points corresponding to the unsafe poses fall outside of the collision-free sphere~\sthree for both Scenarios~1 and~2 (refer to Table~\mref{tb:testcase})}
	\end{figure}
	Since the purpose of this work is only to {\em validate} the value of~$r_3$ already computed using the analytical process (and not to compute it {\em ab initio}), the knowledge of~$r_3$ is utilised explicitly in the process. Thus, a \emph{spherical shell} (denoted by~\sphrf) centred at~$\bp_0$, is generated {\em enclosing} the sphere~\sthree of radius~$r_3$. The inner and outer radii of~\sphrf are given by~$\tilminri = r_3(1-\delta)$ and~$\tilmaxri = r_3(1+\delta)$, respectively. The validation of the safe radius~$r_3$ is done {\em only} within this shell, as opposed to a sphere or some other region of larger dimensions containing the safe sphere~$\mathcal{S}_3$. Obviously, this choice reduces computational expenses by focussing only on the region of interest. In this work,~$\delta$ was chosen to be~$0.1$.

	A set of~$N$ points is generated within~\sphrf (see Step~3 in Section~\mref{sc:api}). Towards this, firstly, a total of~\nr spheres are generated within~\sphrf, by sampling the range~$\left[\tilminri,\tilmaxri\right]$ \emph{uniformly} with a suitable resolution (denoted by~$\Del r$); i.e.,~$r = \tilminri + m\Del r$, where ~$\ m = 0, \dots, \nr$, $\nr = \big\lfloor(\tilmaxri-\tilminri)/\Del r \big\rceil$.
	Then,~\nso samples are generated on the surface of each sphere,  using the~\emph{uniform sampling based on regular placement}~(USRP) method (refer to~\mcite{Patra2023} for the details).
	Thus, a set of~$N = \nr \times \ns$ samples is generated within~\sphrf, each representing a distinct position of the centre of the MP. For each such point in~\sphrf, all the legs are checked for collisions in a pairwise manner. To harness the built-in capabilities of geometric computations of~\invent to perform these checks, the capsules enclosing each leg need to be (logically) decomposed into their constituent elements, namely, a set consisting of two terminal spheres and a (finite) cylinder between them. These geometric primitives are recognised by \invent, while the capsule, in its entirety, is not. Thus, the problem reduces to determining if any of the three elements of such a set (representing one leg) interferes with any other of a similar set (representing another leg). The built-in routine \textsf{AnalyzeInterference} is pressed into action for this purpose. Finally, points in~\sphrf at which no such collision is detected are termed as {\em safe}, and the rest are stored for further analysis to compute the details of the interference. If all the points inside~\cfs are safe, the validation is considered to be successful. 

	\section{Results of numerical experiments}
	\mlabel{sc:result}
	\begin{figure}[t]
		\centering
		\includegraphics[width=\textwidth]{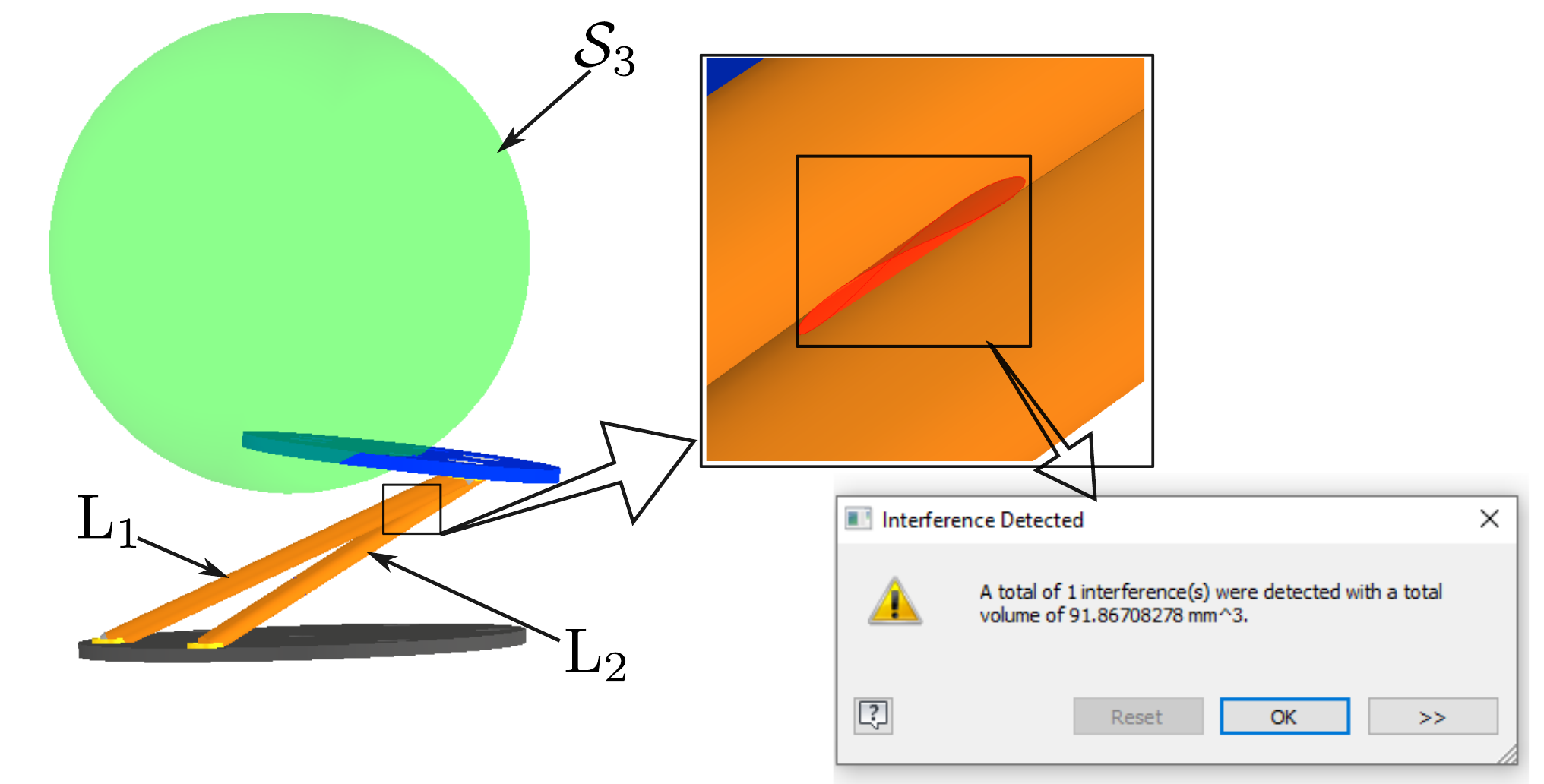}
		\caption{Interference among the legs~$\text{L}_1$ and~$\text{L}_2$ in Scenario~1}
		\mlabel{fg:intl1l2}
	\end{figure}
	\begin{figure}[t]
		\centering
		\includegraphics[width=\textwidth]{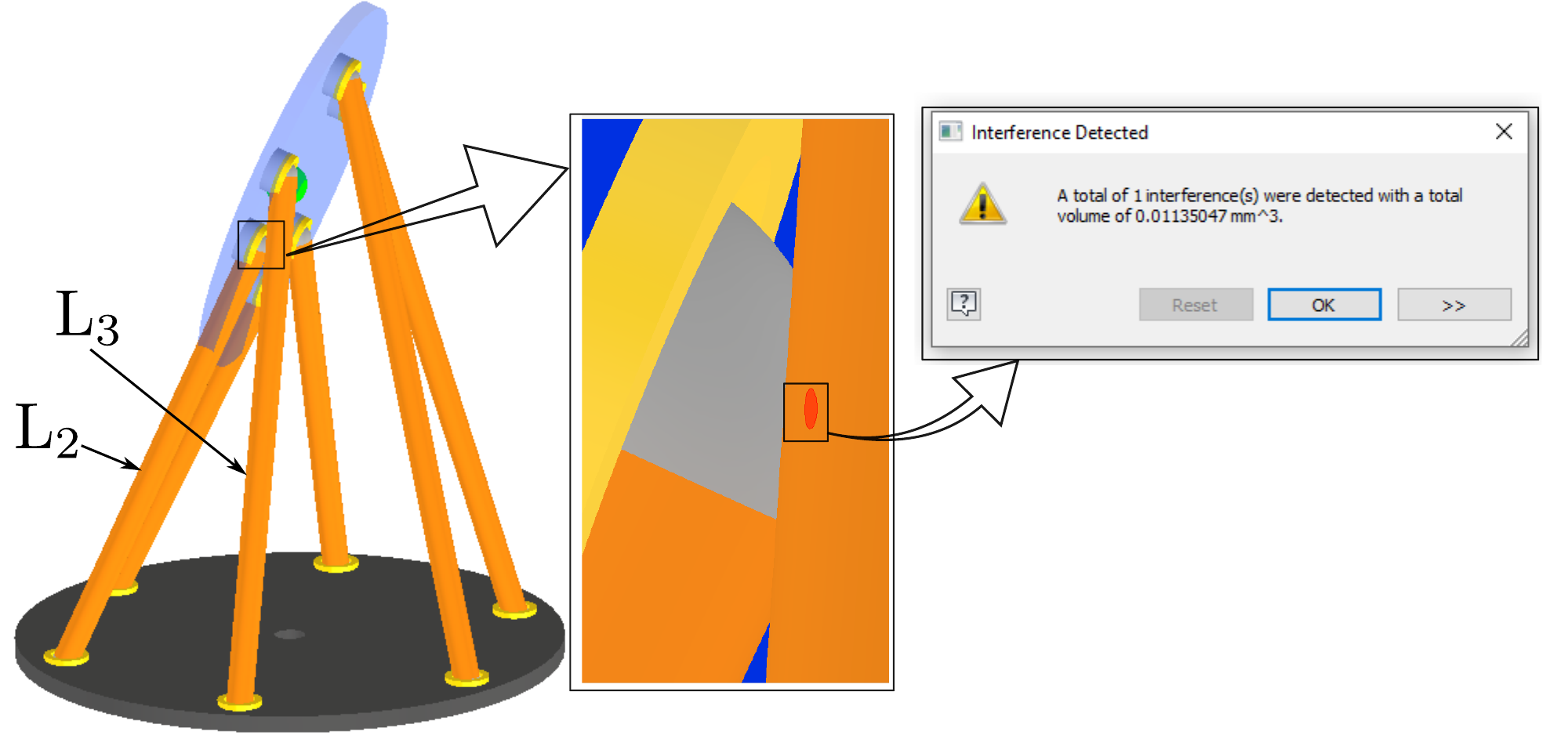}
		\caption{Interference among the legs~$\text{L}_2$ and~$\text{L}_3$ in Scenario~2}
		\mlabel{fg:intl2l3}
	\end{figure}
	The computational process presented above is illustrated in this section via application to a~\cfs that has been computed in~\mcite{PatraS32024}.
	The architecture parameters of the~\srs, as well as the neutral height adopted in~\mcite{PatraS32024}, are listed in Table~\mref{tb:archparam}. Two scenarios are considered in this section: (1)~the radius~$r_3$ of the~\cfs validated considering only one pair of legs, namely,~$\{\text{L}_1,\text{L}_2\}$, and (2)~the radius~$r_3$ of the~\cfs validated considering \emph{all} 15~pairs of legs.
	The Rodrigues parameters of the MP,~\bc, and the corresponding radius~$r_3$ for the above two scenarios are listed in Table~\mref{tb:testcase}.
	All computations are performed on a single thread of an  AMD$^\circledR$ Ryzen$^\text{TM}$ Threadripper~2990WX CPU running at 3.0~GHz.  
	
	\subsection{Validation of the CFS corresponding to Scenarios 1 and 2}
	\mlabel{sc:validcase12}
	In both scenarios, the~\cfs is validated by generating~$N$ points within the associated~\sphrf, 
	whose dimensions and the corresponding resolutions are also specified in Table~\mref{tb:paramscan}. For both scenarios, the number of samples on any sphere within~\sphrf is chosen as~$\ns = 2500$. With these settings, the total time taken to validate the~\cfs for each scenario is approximately 30~minutes.
	
	For all the points within~\sphrf, the unsafe poses are identified following Section~\mref{sc:validcfs}. It is found that in Scenario~1, out of~10,000 points, only~$81$ points are unsafe, wherein out of~7,500 points,~$38$ points are found to be so in Scenario~2. 
	As shown in Figs.~\mref{fg:unsafecase1},~\mref{fg:unsafecase2}, respectively, all the unsafe points fall outside~\sthree in either scenario. Therefore, the~CFSs under investigation are found to be safe in both cases. 
	
	For the sake of completeness, an example of actual detection of collision within Scenario~1 is shown in Fig.~\mref{fg:intl1l2},  wherein the cylinders corresponding to the legs~$\text{L}_1$ and~$\text{L}_2$ interfere at one of the unsafe points shown in Fig.~\mref{fg:unsafecase1}. A complementary case of one of the terminal spheres interfering with the cylinder of another leg is presented in Fig.~\mref{fg:intl2l3}. 
	\section{Conclusions}
	\mlabel{sc:concl}
	This paper presents a computational scheme for validating the largest collision-free sphere for a given orientation of the MP of an~\srs in an automated manner using the~\api of~\invent. The legs of the manipulator are approximated by capsules to facilitate the geometric computations. A number of positions of the MP are generated in the neighbourhood of the boundary of the~\cfs to be validated, and each of them is checked in a sequence for possible collisions between each pair of legs. The results obtained indicate that none of the points lying inside the CFSs under study in two different scenarios lead to any collision; i.e., the computed CFSs are indeed free of collisions (at least as the resolution adopted). Beyond validating pre-computed CFSs, the proposed method can be employed to estimate the dimensions of such CFSs with reasonable accuracy.
	\bibliographystyle{spmpsci} 
	\bibliography{reference}
\end{document}